\documentclass[]{llncs}
\usepackage{graphicx}

\usepackage{epstopdf}
\newcommand{\ie}{\emph{i.e.,}}

    \makeatletter
\def\@fnsymbol#1{\ensuremath{\ifcase#1\or *\or \dagger\or \ddagger\or
   \mathsection\or \mathparagraph\or \|\or **\or \dagger\dagger
   \or \ddagger\ddagger \else\@ctrerr\fi}}
    \makeatother  
\begin{document}
\title{Deep Instance-Level Hard Negative Mining Model for Histopathology Images}
%
%

\author{Meng Li\thanks{\email{meng.li@uq.edu.au}; \email{lin.wu@uq.edu.au}; \email{arnold.wiliem@ieee.org}; \email{\{k.zhao1,patrick.zhang\}@uq.edu.au}; \email{lovell@itee.uq.edu.au}}  \and
Lin Wu\thanks{Corresponding author} \and
Arnold Wiliem \and
Kun Zhao \and
Teng Zhang \and
Brian Lovell}
%
%
\institute{The University of Queensland, School of ITEE, QLD 4072, Australia \\
June 25, 2019
}
\maketitle              
\begin{abstract}
Histopathology image analysis can be considered as a Multiple instance learning (MIL) problem, where the whole slide histopathology image (WSI) is regarded as a bag of instances (\ie\ patches) and the task is to predict a single class label to the WSI. However, in many real-life applications such as computational pathology, discovering the key instances that trigger the bag label is of great interest because it provides reasons for the decision made by the system. In this paper, we propose a deep convolutional neural network (CNN) model that addresses the primary task of a bag classification on a histopathology image and also learns to identify the response of each instance to provide interpretable results to the final prediction. We incorporate the attention mechanism into the proposed model to operate the transformation of instances and learn attention weights to allow us to find key patches. To perform a balanced training, we introduce adaptive weighing in each training bag to explicitly adjust the weight distribution in order to concentrate more on the contribution of hard samples. Based on the learned attention weights, we further develop a solution to boost the classification performance by generating the bags with hard negative instances. We conduct extensive experiments on colon and breast cancer histopathology data and show that our framework achieves state-of-the-art performance.
\end{abstract}
\section{Introduction}
Deep learning has become increasingly popular in the medical imaging area. However, due to high computational cost, working on whole slide histopathology images (WSIs) with gigapixel resolution is challenging. As a consequence, some approaches attempt to divide WSIs into small patches~\cite{2015arXiv150407947H}, and predict the final diagnosis of a WSI without providing the pixel-level annotations. To alleviate the annotation efforts, multiple instance learning (MIL) is introduced and explored for weakly annotated WSIs~\cite{xu2014deep}, where a WSI is considered as a bag, and the patches of that WSI are regarded as instances. Hence, the problem is cast as dealing with a bag of instances for which a single class label is assigned to the WSI. However, providing insights into the contribution of each instance to the bag label is crucial to medical diagnosis. Furthermore, if the features of the negative instances are not well learned, the model will be prone to predict a bag with many negative instances to be a positive bag because of the bias issue~\cite{NIPS2010_3941}. In Liu et al.'s work~\cite{pmlr-v25-liu12b}, the concept of key instances is discussed, which indicates that in the MIL task, discovering the key instances that trigger the bag label is critical. Thus, one reason for a model to make a wrong prediction about a negative bag is that some challenging instances mislead the model. This will affect decision making of a medical diagnosis model, as a single challenging instance in a WSI may lead to a wrong diagnosis result. The challenging instances herein are referred to as \textbf{hard negative patches (instances)} which are incorrectly classified as positive by the prediction model. The other prominent challenge in training a deep model for WSIs is the training imbalance, which means the model tends to become biased towards learning from limited negative patches together with many positive patches.

\begin{figure}[t]
\includegraphics[width=\textwidth]{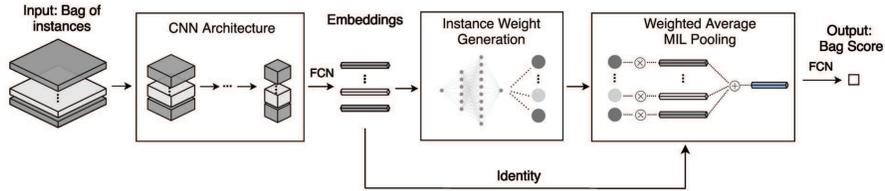}
\caption{\small The architecture of the end-to-end deep CNN model with adaptive attention mechanism. The input is the bag of instances (patches of each WSI), which are fed into a CNN model to produce the latent representation of each instance. Then the embeddings of instances go through a fully connected network with the attention weights generated. The learned weights are multiplied with the embeddings of instances in element-wise to be classified by the classifier.} \label{fig:framework}
\end{figure}

To address these challenges, we propose an approach that generates the hard negative instances in histopathology images based on instance-level adaptive weighting. Specifically, we incorporate the attention mechanism into the training model, which is computed with adaptive weight loss to explicitly assign larger weights to hard negative instances. The selected instances are used to constitute hard training bags through a bag generation algorithm to further improve the training accuracy. The overview of our framework is shown in Fig. \ref{fig:framework}. The major contributions of this paper are summarised below.
\vspace{-0.2cm}
\begin{itemize}
\item We propose a MIL model that incorporates adaptive attention mechanism into an end-to-end deep CNN to detect key instances for histopathology image analysis. 
\item We present a strategy that is able to address the training imbalance issue through adaptive weighting distribution on negative samples.
\item We develop a bag generation algorithm based on the selected hard negative instances and compose the hard bags for the improved classification training.
\item Extensive experiments are conducted on real datasets. Experimental results show that our method improves the accuracy, significantly minimizes false positive rate, and also achieves state-of-the-art performance.
\end{itemize}

\section{Related Work}
\subsubsection{MIL on Histopathology Images}
Different MIL approaches were proposed to work on medical images~\cite{2015arXiv150407947H,10.1007/978-3-319-10470-6_29,kraus2016classifying,couture2018multiple}. A two-stage Expectation Maximization based algorithm combined with a deep convolutional neural network (CNN) works well to classify instances on multiple medical datasets~\cite{2015arXiv150407947H}. In Kandemir et al.'s work~\cite{10.1007/978-3-319-10470-6_29}, a Gaussian process with relational learning is introduced to exploit the similarity between instances of Barrett’s cancer dataset. To relate the instances and the bags, different permutation-invariant pooling approaches with CNN have been proposed. One solution is noisy-and pooling function with CNN, which has achieved promising results on medical images~\cite{kraus2016classifying}. Heather D. Couture et al.~\cite{couture2018multiple} adopt an instance-based approach and aggregate the predictions of patches by using a quantile function on breast cancer. In this paper, our work aims to augment the CNN training with attention mechanism to attentively select hard samples, which achieves state-of-the-art performance on colon cancer and breast cancer dataset~\cite{7399414,10.1007/978-3-319-10470-6_29}.

\subsubsection{Instance-Level Weights}
The attention mechanism has become well known in deep learning, especially for natural language processing tasks and image representation learning~\cite{Deep-recursive,Wu-TMM,CYC-DGH,Wu-TCYB,Deep-Embed}. For MIL problems, attention is represented by the weights of each instance in a bag, a higher weight means more attention. The method to obtain instance weights was first proposed in Nikolaos Pappas and Andrei Popescu-Beliss's work~\cite{pappas2014explaining}, where the weights of instances were trained in a linear regression model. They improved the previous method by using a single layer neural network~\cite{pappas2017explicit}. A more recent solution was introduced, where the weights were learned by a two-layer neural network~\cite{pmlr-v80-ilse18a}. However, these works did not further analyze the generated attention weights. In this work, we propose a method that exploits the relation among the weights and reuses them to improve the model performance. 

\subsubsection{Hard Negative Mining}
The main idea of hard negative mining is to repeatedly bootstrap negative examples by selecting false positives which the detector incorrectly classifies~\cite{dalal2005histograms}. Hard negative mining is originally used in object detection tasks, where the datasets usually involve overwhelming easy examples~\cite{Shrivastava_2016_CVPR}. Recently, this simple but effective technique is commonly used in the medical domain. Difficult negative regions are extracted from the training set to boost model performance on lymph nodes~\cite{wang2016deep} and breast cancer WSIs~\cite{bejnordi2017deep}. Whilst these works make advances in this domain, to our knowledge, there is no work which addresses hard negative mining for MIL tasks. In this paper, we propose an approach which is able to detect hard negative instances by utilizing the attention weight given for each instance in a bag. Based on this, we propose several hard negative bag generation algorithms. As shown in the experiment, our proposed approach achieves significant improvement in comparison with the baselines.

\section{Our Method}
Given a histology image (bag) \( \bf X_{i}\) with its label \(C \in \left\{0,1\right\}\), and its instances \(X_{ij}\in{\bf X_{i}}\), \(j=1,2,...,N_{i}\), our goal is to identify a set of hard negative instances from negative bags \(\bf X_{i}\) with \(C_i=0\). We define \( \bf \hat{X}_{i}= \left\{h_{1},h_{2},...,h_{M}\right\}\) as the set of hard negative instances identified from the negative bags. We generate hard negative bags from \(\bf \hat{X}_{i}\) and use these bags in the training step to improve accuracy. To this end, we develop an instance-level, attention-based CNN model which determines the instance weights by inspecting their corresponding response to contribute the final prediction outcome. The attention selection of key instances allows us to detect the hard negative instances leading to false positives. Then, we propose a bag generation algorithm which produces new bags with selected hard negative patches for retraining.

\subsubsection{Instance-Level Adaptive Weighing Attention}
We implement a neural network \(f_{\phi}\left(\cdot\right)\) to transform the \(j\)-th instance in the \(i\)-th bag into a low-dimensional embedding \(g_{ij} = f_{\phi}\left(X_{ij} \right)\), where \(g_{ij} \in \rm I\!R^{M}\). To get the bag representation which is permutation-invariant, a max or a mean aggregation is usually used in the MIL problem~\cite{kraus2016classifying}. However, neither of them can be trained to get the instance weights which are needed in our case. As a result, a weighted average aggregation method is used. Given the embedding-based instances \(G_{i} = \left\{g_{i1},g_{i2},...,g_{iN_{i}}\right\}\), the weighted average is shown as follows: 
\begin{equation} \label{eq:1}
z_{i} = \sum_{j=1}^{N_{i}}w_{ij}g_{ij}, \quad w_{ij} = \frac{\exp(\mathbf{v}_{i}^\top\frac{\mathbf{U}_{i}{g_{ij}}^\top}{1+\mid{\mathbf{U}_{i}{g_{ij}}^\top}\mid})}{\sum_{j=1}^{N_{i}}\exp(\mathbf{v}_{i}^\top\frac{\mathbf{U}_{i}{g_{ij}}^\top}{1+\mid{\mathbf{U}_{i}{g_{ij}}^\top}\mid})}
\end{equation}
In (\ref{eq:1}), \(\mathbf{v}_{i} \in \rm I\!R^{L\times 1}\) and \(\mathbf{U}_{i} \in \rm I\!R^{L\times M}\) are the attention network weight parameters. A softsign activation function is implemented to ensure proper gradient flow. We normalize the weights \(w_{ij}\) through a softmax function so that all the weights sum to 1. However, the weights defined in (\ref{eq:1}) are not constrained and could make all the instances receive uniformly distributed weights, which would make it difficult to identify hard negative instances. To this end, we introduce an adaptive weighting method to explicitly enlarge the weights difference between the positive and negative instances:
\begin{equation}
z_i=\sum_{j=1}^{N_{i}-N_{in}}w_{ij}g_{ij} + \lambda \sum_{j=1}^{N_{in}}w_{ij}g_{ij},
\end{equation}
where $N_{in}=|\{X_{ij}| j\in pseudo ~negative ~instances\}|$ is the number of pseudo negative instances. In each iteration, the instances are treated as pseudo negative by thresholding on the average weights in each cycle. The balancing hyper-parameter $\lambda$ is empirically set to 2 in our paper. We note that when $\lambda$ is set $> 1$, the weights of the pseudo negative instances will be reduced. This will increase the differences between the positive instance and the negative instances. After obtaining the weighted embeddings, we use a CNN layer for classification. In the end, the weights of instances are learned so that we can make use of them to mine the hard negative instances. 
\vspace{0.1cm}

\begin{figure}[t]
\includegraphics[height=5cm,width=0.9\textwidth]{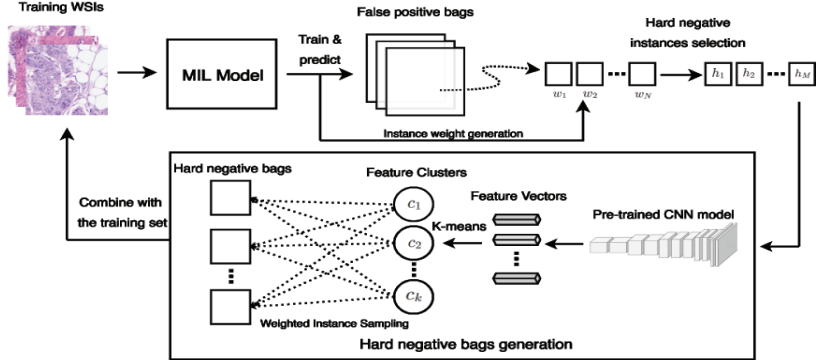}
\caption{\small The proposed novel hard negative mining process. The training images are first fed into the deep MIL model with balanced training to select instances to constitute the false positive bags. We learn attention weights for instances, which can be used to select the hard instances that fool the model to make the wrong prediction. Next, the hard negative instances are grouped to form the new hard negative bags by a bag generation algorithm. The patches firstly go through a pre-trained CNN model without the last layer to produce the feature vectors. A K-means clustering is then used to group features into clusters from which hard negative bags are produced by randomly sampling instances across feature clusters dynamically. Finally, the training is augmented with hard negative bags for improved accuracy.} \label{fig:instance-attention}
\end{figure}

\subsubsection{Hard Negative Instance Mining}
After the training process finishes, we obtain the false positive bags \( \left\{B_{1},B_{2},...,B_{N}\right\}\), where each bag \(B_l\) has the corresponding weights of instances as \( \left\{w_{l1},w_{l2},...,w_{lN_{l}}\right\}\), via selecting the hard negative instances through attention weights: 
\begin{equation} \label{eq:3}
H_{l} = \left\{ w_{li} \mid w_{li} \geq \sigma_{l}+\overline{w_{l}}\right\}
\end{equation}
where \(\sigma_{l}\) is the standard deviation and \(\overline{w_{l}}\) is the mean of the weights in the \(l\)-th bag. We group all hard negative instances from all false positive bags together and obtain the bag \(\hat{X}_i=\left\{\bf h_{1}, \bf h_{2},..., \bf h_{M}\right\}\).

\vspace{0.0cm}
\subsubsection{Hard Negative Bag Generation}
In this stage, new hard negative bags are generated for re-training the network to further improve the bag prediction accuracy. As histology images contain a variety of patterns in negative regions~\cite{6247772}, we propose to select diverse patches into each negative bag such that the deep model will learn features comprehensively. The overview of the bag generation process is shown in Fig. \ref{fig:instance-attention}. The hard negative instances first go through a pre-trained CNN model to convert to feature vectors. The features are then clustered by the K-means algorithm into \(c\) feature groups \(C_1,C_2,...,C_c\), which contain \(N_1,N_2,...,N_c\) items. To conform to the training set bag size, we empirically generate the \(i\)-th new bag \(B_i\) which has a Gaussian random size with \(\sigma\) and \(\mu\), where \(\sigma\) is the standard deviation and \(\mu\) is the mean of all training bag sizes. We limit the size from \(Z_{min}\) to \(Z_{max}\)  where \(Z_{min}\) and \(Z_{max}\) are the minimum and maximum bag size respectively of the training set. We follow a weighted sampling strategy that randomly selects instances from each cluster and puts in \(B_i\). The possibility to designate a cluster to sample instance is denoted as \(P_j = \frac{N_j}{\sum_{j=1}^cN_j}\).

\section{Experiments}
\subsubsection{Datasets}
We conduct experiments on two public datasets: Colon Cancer dataset ~\cite{7399414} and UCSB dataset~\cite{10.1007/978-3-319-10470-6_29}. Colon Cancer dataset involves 100 Hematoxylin and Eosin stained (H\&E) images from 9 patients at 0.55$\mu$m/pixel resolution. The histology images include multiple tissue appearance that belongs to both normal and malignant regions, and can be either used for detection or classification. In total, 22,444 nuclei are annotated in four classes, \textit{i.e.} epithelial, inflammatory, fibroblast, and miscellaneous. A histology image is positive if it consists of at least one epithelial nucleus. We divide each image into 27 \(\times\) 27 patches. The UCSB dataset contains 58 H\&E stained image excerpts (26 malignant, 32 benign) from breast cancer patients. The histology image size is 896 \(\times\) 768 and we divide each image into 32 \(\times\) 32 patches. To reduce the noisy patches, each image is converted from the RGB color space to the HSV space, and an Otsu algorithm is used to select the optimal threshold values in each channel to filter the patches~\cite{otsu1979threshold}. The patches are randomly flipped and rotated for augmentation. Finally, the color normalization (histogram equalization) is performed on each patch.

\subsubsection{Implementation Details}
We adopt a deep CNN architecture~\cite{7399414} and a MIL pooling layer~\cite{pmlr-v80-ilse18a} to extract features and a fully connected layer to make the classification. More detailed treatment for the architecture can be found in the supplementary material. We follow the work~\cite{pmlr-v80-ilse18a}, and train the network using an Adam optimizer with \(\beta_{1}=0.9\) and \(\beta_{2}=0.999\). A cross-entropy loss function is used for regression. For the Colon Cancer/UCSB dataset, we set the learning rate = \(5\times10^{-5}\)/\(5 \times 10^{-6}\) and weight decay = \(5\times10^{-4}\)/\(1\times10^{-4}\). Each experiment runs for 120/300 epochs, and the epoch with the lowest loss is chosen for evaluation. We evaluate our method using 10-fold/4-fold cross-validation and five repetitions for each experiment. To ensure fair comparison, we test on the same set of bags in each fold for the methods.

\begin{table}\scriptsize
\centering
\caption{\small Evaluation of different methods for colon cancer data. The average of five experiments with its corresponding standard error is reported. SB: single bag hard negative mining, MB: multiple bags with random sampling, FMB: multiple bags with features clustering sampling, AUC: area under the curve, FPR: false positive rate.}\label{tab2}
\begin{tabular}{l*6c}
\hline
Method & Accuracy & Precision & Recall & F-score & AUC & FPR\\
\hline
MIL Model~\cite{pmlr-v80-ilse18a} & 0.904\(\pm\)0.011 & 0.953\(\pm\)0.014 & 0.855\(\pm\)0.017 & 0.901\(\pm\)0.011 & 0.968\(\pm\)0.009 & NA\\
Our Model & 0.906\(\pm\)0.007 & 0.912\(\pm\)0.010 & 0.916\(\pm\)0.012 & 0.905\(\pm\)0.008 & 0.952\(\pm\)0.012 & 0.104\(\pm\)0.012\\
Our Model+SB & 0.922\(\pm\)0.004 & 0.937\(\pm\)0.011 & 0.920\(\pm\)0.014 & 0.920\(\pm\)0.005 & 0.979\(\pm\)0.004 & 0.084\(\pm\)0.012\\
Our Model+MB & 0.942\(\pm\)0.005 & 0.963\(\pm\)0.010 & \textbf{0.928}\(\pm\)0.017 & 0.939\(\pm\)0.006 & 0.982\(\pm\)0.009 & 0.052\(\pm\)0.010\\
Our Model+FMB & {\textbf{0.948}}\(\pm\)0.004 & \textbf{0.980}\(\pm\)0.003 & \textbf{0.920}\(\pm\)0.006 & \textbf{0.945}\(\pm\)0.004 & \textbf{0.983}\(\pm\)0.004 & \textbf{0.036}\(\pm\)0.007\\
\hline
\end{tabular}
\end{table}

\begin{figure}[hbt]
\centering
\includegraphics[width=0.85\textwidth]{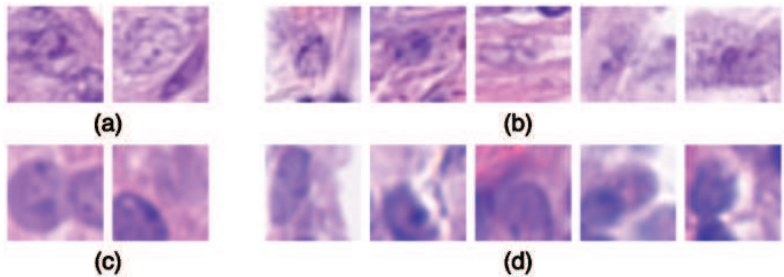}
\caption{Positive and hard negative examples: (a) Colon data: instances that include malignant regions. (b) Colon data: detected hard negative instances that mislead the model to predict a normal bag into a malignant result. (c) UCSB data: instances that include malignant regions. (d) UCSB data: detected hard negative instances that mislead the model to predict a normal bag into a malignant result.} \label{fig:hardnegative}
\end{figure}

\begin{table}[t]\scriptsize
\centering
\caption{\small Evaluation of different methods for the UCSB dataset. The average of five experiments with its corresponding standard error is reported. SB: single bag hard negative mining, MB: multiple bags with random sampling, FMB: multiple bags with features clustering sampling, AUC: area under the curve, FPR: false positive rate.}\label{tab3}
\begin{tabular}{l*6c}
\hline
Method &  Accuracy & Precision & Recall & F-score & AUC & FPR\\
\hline
MIL Model~\cite{pmlr-v80-ilse18a} & 0.755\(\pm\)0.016 & 0.728\(\pm\)0.016 & 0.731\(\pm\)0.042 & 0.725\(\pm\)0.023 & 0.799\(\pm\)0.020 & NA\\
SDR+SVM~\cite{7886294} & \textbf{0.983} & NA & NA & NA & \textbf{0.999} & NA\\
Our Model & 0.821\(\pm\)0.006 & 0.870\(\pm\)0.032 & 0.800\(\pm\)0.038 & 0.806\(\pm\)0.011 & 0.942\(\pm\)0.008 & 0.220\(\pm\)0.058\\
Our Model+SB & 0.893\(\pm\)0.001 & 0.920\(\pm\)0.012 & 0.886\(\pm\)0.024 & 0.887\(\pm\)0.015 & 0.949\(\pm\)0.006 & 0.140\(\pm\)0.019\\
Our Model+MB & 0.936\(\pm\)0.004 & 0.936\(\pm\)0.011 & 0.943\(\pm\)0.014 & 0.935\(\pm\)0.005 & 0.967\(\pm\)0.004 & 0.100\(\pm\)0.016\\
Our Model+FMB & \textbf{0.975}\(\pm\)0.008 & \textbf{0.975}\(\pm\)0.012 & \textbf{0.979}\(\pm\)0.009 & \textbf{0.975}\(\pm\)0.007 & \textbf{0.999}\(\pm\)0.001 & \textbf{0.040}\(\pm\)0.019\\
\hline
\end{tabular}
\end{table}


\subsubsection{Results and Discussion}
Table 1 and Table 2 show the results of our method against different baselines on two datasets. We compare different hard negative mining strategies, \textit{i.e.}, the single hard negative bag generation (SB), the randomly generated multiple bags (MB), and the features clustering bags (FMB). Among these methods, the proposed FMB solution has an overall better result than the others and sets the new state-of-the-art result for colon cancer dataset. For the UCSB dataset, We achieve the same area under the curve (AUC) result as the state-of-the-art method in Song et al.'s work~\cite{7886294}. Fig. \ref{fig:hardnegative} demonstrates the detected hard negative instances by our solution. It is noticeable that compared to the baseline method, the false positive rate (FPR) is significantly decreased, and the recall is increased by including hard negative mining, which is particularly essential in histopathology image diagnosis because both false positive and false negative could result in severe consequences for patients. It is also critical that the AUC is increased, as this indicates that the generated hard negative bags do not cause class imbalance issues.

\vspace{0.0cm}
\section{Conclusions}
In this paper, we introduce an effective approach to MIL tasks on histopathology data that incorporates the attention mechanism with adaptive weighing into the deep CNNs for balanced training. To further improve the training accuracy, we develop a novel hard negative mining strategy by generating the bags with hard negative instances. Experimental results demonstrate that our approach makes a decent improvement from the baseline method and achieves state-of-the-art performance. In future work, we plan to evaluate our model on the larger WSIs. Moreover, we intend to generalize our model to the multi-class task not limited in the binary classification. 

\subsubsection{Acknowledgement}
This research was funded by the Australian Government through the Australian Research Council and Sullivan Nicolaides Pathology under Linkage Project LP160101797.

\begin{scriptsize}
%
%
%
%

\end{scriptsize}

\end{document}